# Harmonizing Metadata of Language Resources for Enhanced Querying and Accessibility

Zixuan Liang
Harrisburg University of Science & Technology
zliang1@my.harrisburgu.edu

*Abstract*—This paper addresses the harmonization of metadata from diverse repositories of language resources (LRs). Leveraging linked data and RDF techniques, we integrate data from multiple sources into a unified model based on DCAT and META-SHARE OWL ontology. Our methodology supports text-based search, faceted browsing, and advanced SPARQL queries through Linghub, a newly developed portal. Real user queries from the Corpora Mailing List (CML) were evaluated to assess Linghub's capability to satisfy actual user needs. Results indicate that while some limitations persist, many user requests can be successfully addressed. The study highlights significant metadata issues and advocates for adherence to open vocabularies and standards to enhance metadata harmonization. This initial research underscores the importance of API-based access to LRs, promoting machine usability and data subset extraction for specific purposes, paving the way for more efficient and standardized LR utilization.

## I. INTRODUCTION

The study of language and the development of Natural Language Processing (NLP) applications rely on language resources (LRs). Recent initiatives, such as META-SHARE [1], provide digital repositories that index metadata for LRs, supporting their discovery and reuse. META-SHARE offers an open, interoperable infrastructure where LRs can be documented, uploaded, stored, cataloged, exchanged, and discussed. It includes a metadata schema implemented as an XML Schema Definition (XSD), allowing descriptions of LRs throughout their lifecycle. Other repositories, like ELRA [1], LDC [2], CLARIN VLO [2], the Language Grid [3], and Alveo [3], also provide access to LRs, with varying metadata schemes. However, these sources do not offer integrated and uniform querying.

Linked Data offers a solution by linking datasets conceptually, facilitating access and reuse. Applying Linked Data technologies to LRs could create an interconnected ecosystem of resources [4]. Building on this concept, Linghub integrates metadata from various repositories, harmonizing them to offer a unified entry point for discovering language resources across platforms.

Thus, in this paper, we are concerned with how to facilitate the discovery of language resources for a particular task. Given the fact that metadata records for resources are distributed among different catalogs and repositories makes the task of finding a particular resource meeting certain requirements very challenging. This is shown by the fact that many emails to dedicated mailing lists such as the Corpora mailing list contain requests for resources meeting certain desiderata with respect to type of resource (corpus, dictionary, parallel text, etc.), language, size (in tokens or sentences), etc. So far, no repository that allows discovery of resources across repositories has been available. We have closed this gap by developing Linghub, a linked data based portal that indexes and aggregates metadata entries from different repositories including META-SHARE, the CLARIN VLO, LRE-Map and Datahub.io.

Parts of the Linghub as a technical system has been described before [5], [6]. However, this manuscript is the only comprehensive description of the principles behind Linghub, a detailed description of the harmonization procedures and this paper includes results of a novel evaluation. Linghub not only indexes the metadata entries, but also harmonizes the information by mapping it to standard semantic web vocabularies as well as to a recently created ontology of language resources[4] that has been developed on the basis of the existing META-SHARE schema [7]. For this, it relies on state-of-the-art word sense disambiguation methods to support the normalization of data.

One of the crucial technologies that enables this integration is that of RDF [8] and linked data. Linked data is based on four fundamental princi- pals [9].

1) Use Uniform Resource Identifiers (URIs) to identify everything in a resource, thus ensuring that every element of the resource can be identified in a standard manner.
2) Furthermore, use HTTP URIs as they require the association with domain names, ensuring that the data can be clearly traced to its host and thus someone responsible for that dataset.
3) Ensure that URIs resolve, in the sense that when typed into a web browser an appropriate description of the resource is obtained. Ideally the server should detect (using content negotiation) the type of the user and provide HTML results for humans and RDF (serialised in XML or JSON, for instance) for software agents.
4) Provide links to other resources so that it is possible to identify commonalities between resources and to handle

---
[1] http://www.elra.info
[2] https://www.ldc.upenn.edu/
[3] http://alveo.edu.au/
[4] http://purl.org/net/def/metashare

issues of semantic interoperability and provenance.

Linked data makes a highly appropriate model for the task of integrating information about language resources as a Web portal allows use to define stables URIs. we harmonize data by attempting to map metadata descriptions from different sources into the same data schema and enforce the use of the same URIs for equivalent metadata elements as well as values. HTML descriptions are provided at the URIs and in order to meet use cases for automatic training of NLP systems, we find the provision of a machine-readable API also of vital importance. Finally, links to other resources are vital to not only provide links back to the source records, but also to ensure that users can find resources for their needs.

In this paper, besides describing Linghub and the semantic normalization methods used, we provide an evaluation of the ability of Linghub to answer the needs of actual users seeking language resources meeting certain criteria. For this, we have analyzed user requests for resources issued on the Corpora List and analyzed in how far Linghub is able to answer them. This evaluation is the main contribution of the current paper. We are not aware of any similar evaluation conducted in the context of repositories of language resources, so that to our knowledge this is the first attempt to evaluate the ability of a repository to answer requests for language resources.

This paper is structured as follows: Firstly, in Section II we will discuss some of the existing related work in particular focussing on the metadata repositories that we will integrate and in Section III we will describe how we collected the data. In Section IV we derive a single data model based on existing standards that will allow us to combine all the resources and then we will show our procedure for harmonizing these resources. In Section V, we will describe the data portal and provide a thorough evaluation of the system based on real-world queries and thus show the effectiveness of our approach, and finally we conclude in Section VI.[5]

## II. RELATED WORK

Harmonization of data is an important challenge within many application domains. In particular, the integration of data or metadata from different origins is a major challenge, typically requiring a lot of effort. Thus, there is a key interest in developing methods that can support the automatization of this task to a large extent. The domain of language resources is no exception here, with many different metadata schemas and repositories existing that render the integration of all these metadata a very challenging task. Nilsson [10] proposed a framework supporting data integration. He has argued that the integration at a syntactic level can only be a first step towards harmonization and that the holy grail lies on the integration at the semantic level. He proposes RDF as a suitable model for achieving semantic integration.

[5]Some results in this paper have been previously published in the following workshop papers [6], [5], [7]. The results and methods are expanded and combined in this paper in line with the journal guidelines. In addition, Section V provides a novel evaluation of the system.

Khoo and Hall [11] worked on integrating metadata from the Internet Public Library and the Librarian's Internet Index and conclude that the integration of data such data is a very 'resource-intensive' and 'ad-hoc' process. Nogueras et al. [12] similarly developed 'crosswalks' for geographic data and stress the need for formal modelling by way of ontologies for verifying such crosswalks. Chan and Zeng [13] also focus on the use of crosswalks and its use in providing optimal access to data.

A potential solution to data integration problems is in the development of highly reusable data schemas that are reused by many stakeholders. This is for instance the case of the Dublin Core vocabulary [14] that defines a set of properties that can be widely reused across applications. In fact, the development of many small vocabularies that can be widely reused lies at the hear of the Linked Data and Semantic Web initiatives [15]:

> "A larger set of ontologies sufficient for particular purposes should be used instead of a single highly constrained taxonomy of values."

This practice of developing small but highly reusable vocabularies stands in contrast to the practice of many organization in developing large and proprietary data schemas to represent and model their data. This is a legitimate strategy as every organization has a keen interest to have a data schema reflect their own particularities, environment, way of working, etc. On the down-side, such proprietary data schemas make integration of data across source extremely challenging.

Towards larger harmonization, two strategies are thinkable. Each data provider maps a subset of their own proprietary schema to open and reusable vocabularies as described above. Another alternative, pursued in this paper, is that some intermediate entity, Linghub in our case, maps the data from different sources or repositories to open and reusable vocabularies a posteriori. This is exactly the strategy we pursue with Linghub and what we mean by harmonization. In fact, reuse of URIs is a key issue in harmonization as many integration projects have failed due to lack of reuse of identifiers for the same properties [16].

Several metadata vocabularies in the above sense have been proposed to support uniform description of metadata, most notably the VoID model [17], the DCAT model [18] and its recent extension DataID [19].

Adhering to such vocabularies allows for machine processability of the metadata records from different sources and thus eases the task of collecting and integrating metadata from different origins [20].

In the context of language resources, there have been a number of attempts to collect generic metadata about language resource. As a prominent initiative in this line there is META-SHARE [21], [22], which has developed rich XML-based data schemas for the representation of metadata about language resources. Interoperability of these descriptions with other descriptions is, however, low, as META-SHARE adopts the above mentioned monolithic, highly proprietary metadata schema approach. Another approach that relies on institutional collection of metadata was taken by the CLARIN project, whose *Virtual Language Observatory* [2] collects metadata from a number of host institutes by means of OAI-PMH [23] with a small amount of harmonization provided by the *CMDI Component Specification Language* [24]. A similar project called SHACHI [25] has worked on collecting resources on Asian languages. Another approach in this area is the use of International Standardized Language Resource Numbers [26, ISLRN], where basic metadata has been collected about each resource and they are assigned an identifier based on a single number.

As an alternative to the institutional approach, some resources have relied on self-reporting of resource metadata, most notably the LRE-Map [27], which collects information from authors at major research conferences in computational linguistics and as such they are able to collect information on a wide variety of language resources but often leads to quality issues. A similar project, the Open Language Archive Community [22, OLAC] is in between both approaches collecting resources from a wide community but trying to bring them into a very fixed schema for their resources. Finally, we note the work of the *Open Linguistics Working Group* [28], a community which has promoted the use of open data and produced a 'cloud diagram' showing the adoption of linked data language resources over the last four years.

## III. DATA COLLECTION

To provide comprehensive metadata for a large number of language resources, we sourced data from four main repositories that are openly licensed. These are:

**META-SHARE:** A portal created by the META-NET project, offering detailed descriptions of language resources, mostly curated manually. We used the custom framework LIXR to convert its complex XML format into RDF after defining transformation rules manually. An OWL ontology was developed in collaboration with META-SHARE to enhance interoperability, which we also reused in Linghub [21] [7].

**CLARIN VLO:** The Virtual Language Observatory (VLO) by CLARIN contains resources from multiple institutions, with metadata following the CMDI infrastructure [24]. Data is presented in various XML schemas, and we developed export scripts for the top 10 formats, integrating them into Linghub.

**Datahub.io:** A CKAN-based portal focused on open and linked data. While much of its data is unrelated to language resources, we filtered and imported only relevant datasets in RDF using the DCAT vocabulary. This process was straightforward due to the API support.

**LRE-Map:** A resource populated through contributions from various NLP conferences, with LREC-2014 data available under an open license [27]. The integration was challenging due to RDF errors, such as non-resolving URI schemes, that we had to address [29].

We also explored several sources that could not be included in Linghub due to licensing constraints:

**OLAC:** A collection with various XML schemas per data provider, similar to CLARIN, but with non-open licensing.

**ELRA/LDA:** Data from the European Language Resource Association and Linguistic Data Consortium, which we converted to RDF through custom scripts.

The overall size of all resources is summarized in Table I.

TABLE I
THE SIZES OF THE RESOURCES IN TERMS OF NUMBER OF METADATA RECORDS AND TOTAL DATA SIZE

| Source | Records | Triples | Triples per Record |
|---|---|---|---|
| META-SHARE | 2,442 | 464,572 | 190.2 |
| CLARIN | 144,570 | 3,381,736 | 23.4 |
| Datahub.io | 218 | 10,739 | 49.3 |
| LRE-Map (LREC 2014) | 682 | 10,650 | 15.6 |
| LRE-Map (Non-open) | 5,030 | 68,926 | 13.7 |
| OLAC | 217,765 | 2,613,183 | 12.0 |
| ELRA Catalogue | 1,066 | 22,580 | 21.2 |
| LDC Catalogue | 714 | n/a | n/a |

TABLE II
THE RELATIVE NUMBER OF RESOURCES IN EACH OF THE SCHEMAS USED BY CLARIN

| Component Root Tag | Institutes | Frequency |
|---|---|---|
| Song | 1 (MI) | 155,403 |
| Session | 1 (MPI) | 128,673 |
| OLAC-DcmiTerms | 39 | 95,370 |
| MODS | 1 (Utrecht) | 64,632 |
| DcmiTerms | 2 (BeG,HI) | 46,160 |
| SongScan | 1 (MI) | 28,448 |
| media-session-profile | 1 (Munich) | 22,405 |
| SourceScan | 1 (MI) | 21,256 |
| Source | 1 (MI) | 16,519 |
| teiHeader | 2 (BBAW, Copenhagen) | 15,998 |

## IV. MODELLING AND HARMONIZATION OF LANGUAGE RESOURCE METADATA

The foundational model for Linghub is based on the DCAT vocabulary [18], which defines datasets, distributions, and catalogues. However, we extended DCAT to better capture the specific details relevant to language resources by incorporating the META-SHARE ontology [7], which provides richer metadata. For example, META-SHARE includes details such as contact information, versioning, validation, and usage, which were added directly to the model. We also addressed inconsistencies in the application of properties like rights statements by linking nested elements to root data for compatibility with DCAT. Additionally, the META-SHARE ontology models specific elements of language resources, including resource types such as corpora, tools, and lexical resources, along with properties like media type and encoding.

To address the fragmented nature of metadata from different sources, we focused on harmonizing key properties that are critical for resource usage, such as resolution, license type, and the resource's language(s). A major challenge was the variation in resource descriptions across platforms. For example, LRE-Map often uses free-text language names, while META-SHARE employs standardized codes like ISO 639-3. We normalized language identifiers by using ISO 639-3 and applying string similarity metrics like Dice Coefficient and Levenshtein Distance, achieving high accuracy in the harmonization process. Furthermore, resources were categorized according to types, such as 'Corpus', 'Lexical Conceptual Resource', and 'Tool/Service', using the Babelfy linking algorithm [30] to map resources to relevant synsets.

A significant hurdle for resource reusability is the availability of datasets, as many URLs in metadata sources are outdated or do not directly link to data. Our study found that 95% of URLs in the metadata resolved successfully, but the majority led to human-readable pages rather than directly accessible data files. Only 14% of resources were in machine-readable formats like RDF or XML. In addition, the harmonization of license information across platforms was challenging. While some portals, such as META-SHARE, used well-defined URIs for licenses, others relied on free-text fields, complicating machine processing. We addressed this by aligning license information where possible and focused on resources with clear URIs.

Duplicate detection is essential for providing a unified view of resources. We distinguished between intra-repository duplication, where the same resource is reported multiple times within a single repository, and inter-repository duplication, where the same resource is described across different repositories. For example, duplication in META-SHARE was often due to export errors, while in CLARIN, resources like the 'Universal Declaration of Human Rights' had individual pages for each language, which were merged. A sampling of 100 resources revealed a high accuracy rate (98%) in identifying duplicates by title and URL, further refining our resource consolidation process.

TABLE III
THE DISTRIBUTION OF THE 10 MOST USED FORMATS WITHIN THE ANALYZED SAMPLE OF URLS. NOTE XML IS ASSOCIATED WITH TWO MIME TYPES.

| Format | Resources | Percentage |
|---|---|---|
| HTML | 67,419 | 66.2% |
| RDF/XML | 9,940 | 9.8% |
| JPEG Image | 6,599 | 6.5% |
| XML (application) | 5,626 | 5.6% |
| Plain Text | 4,251 | 4.2% |
| PDF | 3,641 | 3.6% |
| XML (text) | 3,212 | 3.2% |
| Zip Archive | 801 | 0.8% |
| PNG Image | 207 | 0.2% |
| gzip Archive | 181 | 0.2% |

## V. LINKED DATA INTERFACE

Linghub provides a web-based portal for accessing harmonized data, as shown in Figure 3[6], offering both human-friendly HTML views and machine-readable formats, including RDF/XML, Turtle, N-Triple, and JSON-LD. The interface is designed for consistent data presentation across sources.

Data discovery is supported by:

[6] http://linghub.org/

TABLE IV
LANGUAGE CONSTRAINTS IN REQUESTS AND RESOURCES AVAILABLE IN LINGHUB COVERING SPECIFIC NUMBERS OF LANGUAGES

| Language count constraint | CML requests | LH resources |
|---|---|---|
| 1 | 8 | 51350 |
| 2 | 1 | 884 |
| 3 | 2 | 69 |
| 4 | 1 | 31 |
| $\geq 5, \leq 19$ | 0 | 54 |
| $> 20$ | 2 | 4 |
| unspecified | 7 | 635895 |

TABLE V
SEARCH RESULTS BY RELEVANCE FOR FREE-TEXT AND SPARQL SEARCH

| | Free Text | | SPARQL | |
|---|---|---|---|---|
| Query | Results | Relevant | Results | Relevant |
| 1 | 88 | 11.36% | 60 | 23.33% |
| 2 | 60 | 10.00% | 63 | 3.17% |
| 3 | 23 | 4.35% | 21 | 4.76% |
| 4 | 1 | 100.00% | 4 | 100.00% |
| 5 | 0 | 0.00% | 16 | 0.00% |
| 7 | 0 | 0.00% | 6 | 16.67% |
| 8 | 73 | 1.37% | 24 | 16.67% |
| 9 | 28 | 14.29% | 11 | 63.64% |
| 10 | 31 | 35.48% | 20 | 100.00% |
| 11 | 47 | 2.13% | 16 | 37.50% |
| 12 | 13 | 46.15% | 4 | 100.00% |
| 13 | 1 | 100.00% | 4 | 25.00% |
| 14 | 18 | 5.56% | 11 | 27.27% |
| 15 | 16 | 6.25% | 2 | 0.00% |
| 16 | 59 | 47.46% | 56 | 48.21% |
| 17 | 4 | 25.00% | 1 | 100.00% |
| 18 | 0 | 0.00% | 0 | 0.00% |
| 19 | 39 | 2.56% | 20 | 0.00% |
| 20 | 81 | 0.00% | 4 | 0.00% |
| 21 | 42 | 7.14% | 5 | 60.00% |
| 22 | 1 | 0.00% | 9 | 0.00% |
| 23 | 14 | 14.29% | 1 | 100.00% |
| Average | 27.78 | 18.84% | 15.57 | 35.92% |

TABLE VI
PORTIONS OF LINGHUB RESOURCES CARRYING AT LEAST ONE PROPERTY VALUE FOR THE RESPECTIVE REQUIRED FACET

| Required Facet | Absolute Freq | Relative Frequency |
|---|---|---|
| (none) | 688287 | 100% |
| Title | 331199 | 48.12% |
| Description | 89053 | 12.94% |
| Language | 52392 | 7.61% |
| Type | 62063 | 9.02% |
| Rights | 36869 | 5.36% |
| Creator | 244725 | 35.56% |
| Subject | 72768 | 10.57% |
| Contact Point | 2436 | 0.35% |
| Access URL | 229020 | 33.27% |

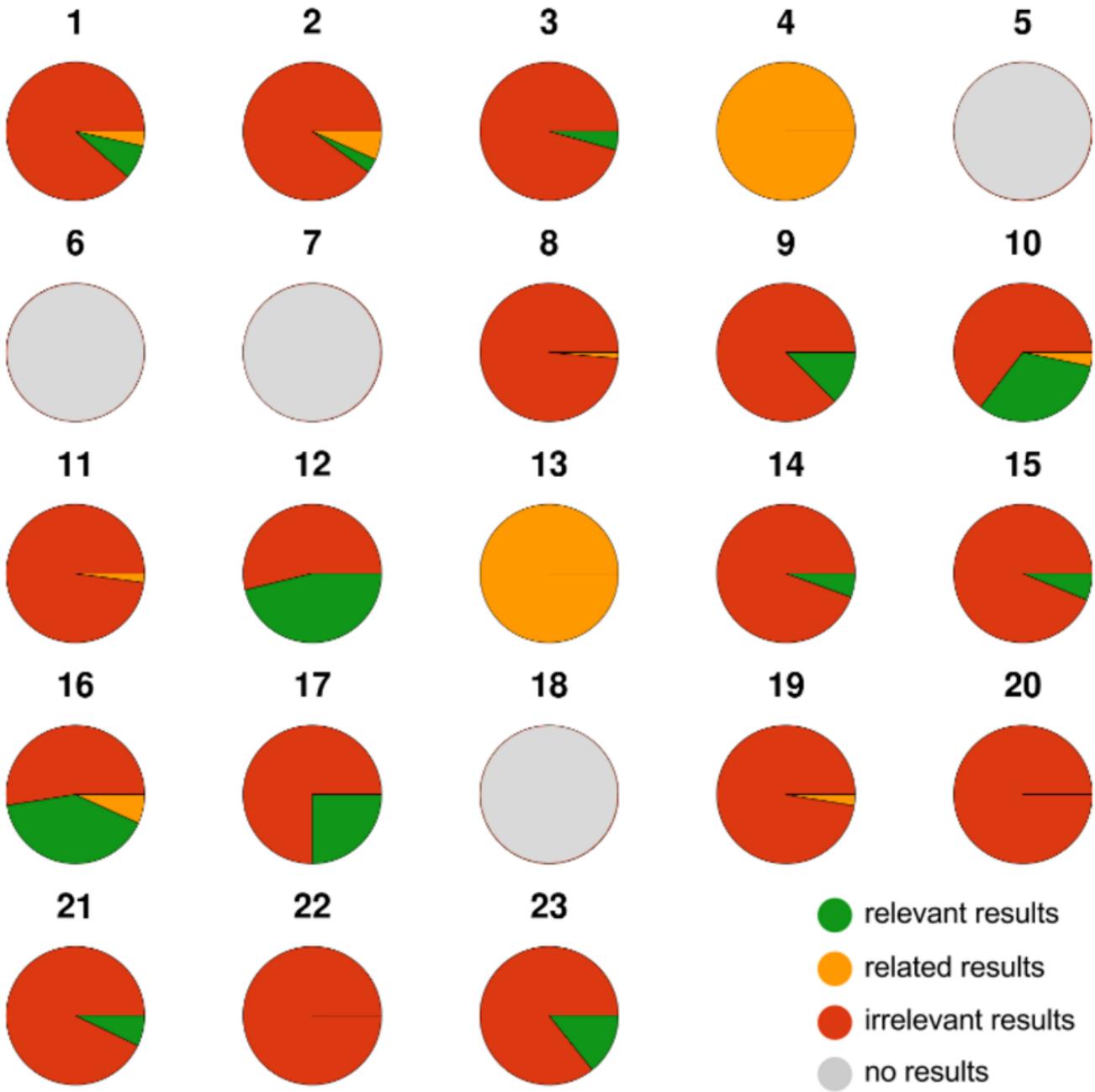

Fig. 1. Percentages of relevant standard interface search results

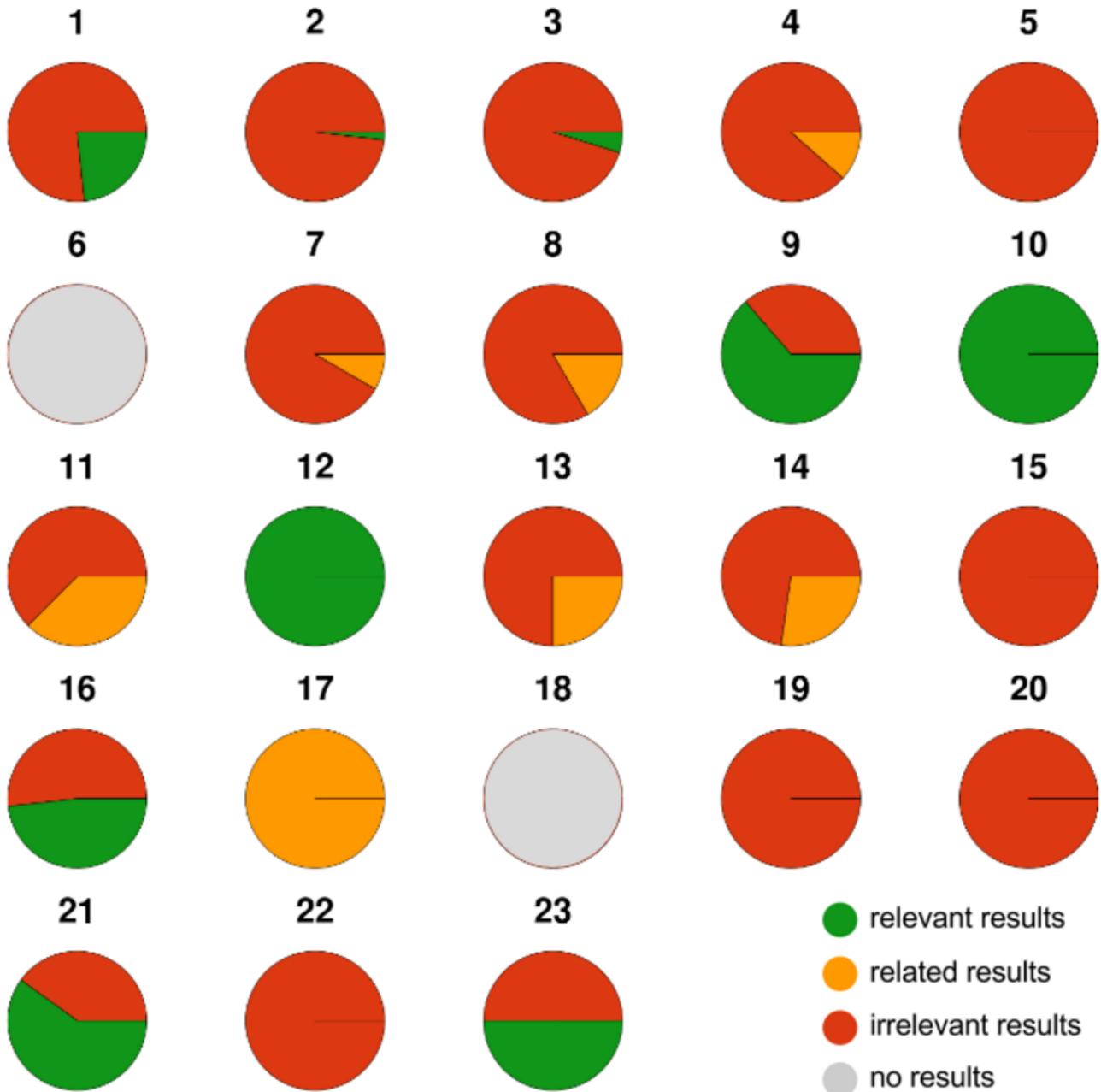

Fig. 2. Percentages of relevant SPARQL search results

**Faceted browsing:** Filters datasets by attributes such as language, rights, and creator.
**Free-text search:** Indexes literal values for search, with additional indexing for languages.
**SPARQL search:** Enables advanced querying, optimized for performance, with results returned in JSON [31].

*A. Evaluation Methodology*

To assess Linghub's usefulness, we analyzed all resource requests posted on the Corpora Mailing List (CML) between January 1st and June 3rd, 2015[7]. After excluding vague or unclear queries, we compiled a set of 23 relevant requests, each containing constraints on resource type, language, extent, and intended use. These were then searched using both the stan-dard search interface and SPARQL queries, accommodating both general users and experts. Results were categorized as relevant, irrelevant, or related, with one request excluded due to ambiguity.

[7]http://mailman.uib.no/public/corpora/

| | | | | | |
|---|---|---|---|---|---|
| **Spanish LMF Apertium Dictionary** | | HTML | RDF/XML | N-Triples | Turtle | JSON-LD |
| *Instance of:* Resource Info | | | | | |

| | |
|---|---|
| Description | This is the LMF version of the Apertium Spanish dictionary. Monolingual dictionaries for Spanish, Catalan, Gallego and Euskera have been generated from the Apertium expanded lexicons of the es-ca (for both Spanish and Catalan) es-gl (for Galician) and eu-es (for Basque). Apertium is a free/open-source machine translation platform, initially aimed at related-language pairs but recently expanded to deal with more divergent language pairs (such as English-Catalan). The platform provides: a language-independent machine translation engine; tools to manage the linguistic data necessary to build a machine translation system for a given language pair and linguistic data for a growing number of language pairs. |
| Language | es |
| Language | Spanish |
| Rights | GPL |
| See Also | http://metashare.elda.org/repository/browse/c19c566292c211e28763000c291ecfc80a823eb7acd74cda8594e986e44407eb/ |

Fig. 3. Screenshot of the Linghub interface

TABLE VII
ACCURACY OF LANGUAGE MAPPINGS

| Resource | Label Accuracy | Instance Accuracy |
|---|---|---|
| SIL *dice coefficient* | 81% | 99.50% |
| SIL *levenshtein* | 72% | 99.42% |
| BabelNet *dice coefficient* | **91%** | 99.87% |
| BabelNet *levenshtein* | **89%** | 99.85% |
| SIL + BabelNet *dice coefficient* | **91%** | 99.87% |
| *levenshtein* | **89%** | 99.85% |

TABLE VIII
THE NUMBER OF INTRA-REPOSITORY DUPLICATE LABELS AND URLS FOR RESOURCES

| Resource | Duplicate Titles | Duplicate URLs |
|---|---|---|
| CLARIN (same contributing institute) | 50,589 | 20 |
| Datahub.io | 0 | 55 |
| META-SHARE | 63 | 967 |
| LRE-Map | 763 | 454 |

*1) Limitations:* Several limitations affect the evaluation. The CML queries are biased, as users likely searched other repositories before posting, making the queries more specific and complex. Most requests were expert-level, limiting the general applicability. Although six months of data were analyzed, the total number of queries was small, and translating the requests into search queries posed additional challenges. Furthermore, the relevance judgments were made subjectively by two annotators, introducing potential variability.

### B. Resource Request Analysis

The resource requests expressed at least two constraints, typically involving resource type and language. Corpora were the most requested resource type (70%), followed by tools (17.4%) and lexical resources (13.0%), aligning with the distribution of resources in Linghub, where 71.5% are corpora. Most requests sought single-language resources, with only a few requesting multiple languages, reflecting the distribution in Linghub's data. Additionally, many requests specified other restrictions, such as corpus size or annotation type, but assessing relevance based on these criteria required subjective evaluation.

### C. Standard Search Interface

Several limitations were identified with Linghub's standard search interface. The inability to apply combined restrictions across multiple facets (e.g., description, language, rights) hindered the expression of more complex queries. Despite this, the full-text search feature in SQLite was helpful for handling "soft" constraints, such as full-text matches in resource descriptions. The interface, however, does not support multiple constraints at once, and it indexes descriptions in multiple languages, though the Linghub interface itself is currently only in English. Of the results, 9.7% were relevant, and 2.6% were related, with many irrelevant results due to the intentionally broad queries aimed at maximizing recall.

### D. SPARQL Search

SPARQL queries, which allow granular search across facets and logical filters, resolved many of the issues with the standard interface. Although SPARQL was more effective, returning 79 relevant results compared to 62 from free-text searches, it remains inaccessible to most linguists due to the need for query language expertise. SPARQL queries were executed on the Linghub dump, and results are available online[8]. Results showed that SPARQL queries were more accurate (22.0% relevant, 8.7% related), demonstrating that the more detailed querying capabilities of SPARQL lead to better results, though constraints like language and size remain challenging.

[8]https://figshare.com/articles/LinghubCMLQueries_pdf/3370969

TABLE IX
NUMBER OF DUPLICATE INTER-REPOSITORY RECORDS BY TYPE

| Resource | Resource | Duplicate Titles | Duplicate URLs | Both |
|---|---|---|---|---|
| CLARIN | CLARIN (other contributing institute) | 1,202 | 2,884 | 0 |
| CLARIN | Datahub.io | 1 | 0 | 0 |
| CLARIN | LRE-Map | 72 | 64 | 0 |
| CLARIN | META-SHARE | 1,204 | 1,228 | 28 |
| Datahub.io | LRE-Map | 59 | 5 | 0 |
| Datahub.io | META-SHARE | 3 | 0 | 0 |
| LRE-Map | META-SHARE | 91 | 51 | 0 |
| All | All | 2,632 | 4,232 | 28 |

TABLE X
PRECISION OF MATCHING STRATEGIES FROM A SAMPLE OF 100

| Duplication | Correct | Unclear | Incorrect |
|---|---|---|---|
| Titles | 86 | 6 | 8 |
| URLs | 95 | 2 | 3 |
| Both | 99 | 1 | 0 |

*E. Data Completeness and Quality*

Linghub's data completeness was evaluated based on the presence of property-value pairs across various facets. Significant variation was found, especially for language and size information. For example, many resources lacked the 'dc:language' property, even when keywords such as "Spanish" appeared in their descriptions. Combining language properties with text-matching in descriptions, as done in SPARQL queries, helped address this. Resources with size data, mostly from META-SHARE, showed that adding structured size information for more corpora would be beneficial. While size can be inferred from descriptions, achieving accurate results would require extensive manual curation.

## VI. CONCLUSION

In this paper, we have addressed the harmonization of metadata about language resources from different repositories. We have motivated why this is an important problem that requires investigation. We have built on linked data and RDF techniques to collect data from several repositories of language resources as a proof-of-concept. We have mapped the data to a data model based on DCAT and the META-SHARE OWL ontology, thus providing the foundation for integration of the data from different repositories at the semantic level. We have develop methods for harmonization and show that for four key descriptive characteristics, that is language, use, type and rights with high accuracy and made the results available through Linghub, a portal that supports different querying mechanisms including a text-based search, faceted browsing functionality as well as advanced search by way of SPARQL. We have collected real user queries from the Corpora Mailing list and evaluated in how far requests for resources can be actually answered by Linghub. While there are obvious limitations, we have shown that many requests can be successfully answered. We are not aware of any other rigorous evaluations of repositories in how far real user needs can be satisfied and are thus the first to provide such an analysis. We consider this as a first solid step towards further research in harmonization of metadata of LRs. As part of our research we have highlighted significant issues that should be considered by metadata providers. Our work would have been simplified if data providers would adhere to open vocabularies and existing standards and identifier systems for representing at least language information as well as right information. Working towards achieving this level of harmonization by reuse of standards is an important endeavor for the LR community that should be jointly address to foster the reuse of LRs beyond the LR community proper. The long-term goal should be to provide API-based for machines to LRs so that machines can automatically decide if a certain language resource is usable for a particular purpose, licensing conditions are favorable. Ultimately, machines should be able to extract subsets of data for a particular purpose, which requires that the data is actually accessible in standard formats. There is a long way ahead of us.

## REFERENCES

[1] C. Federmann, I. Giannopoulou, C. Girardi, O. Hamon, D. Mavroeidis, S. Minutoli, and M. Schröder, "META-SHARE v2: An open network of repositories for language resources including data and tools," in *Proceedings of the 8th International Conference on Language Resources and Evaluation*, 2012, pp. 3300–3303.
[2] D. Van Uytvanck, H. Stehouwer, and L. Lampen, "Semantic metadata mapping in practice: the virtual language observatory," in *Proceedings of the 8th International Conference on Language Resources and Evaluation*, 2012, pp. 1029–1034.
[3] T. Ishida, "Language grid: An infrastructure for intercultural collaboration," in *IEEE/IPSJ Symposium on Applications and the Internet (SAINT 2006)*, 2006, pp. 96–100.
[4] C. Chiarcos, J. McCrae, P. Cimiano, and C. Fellbaum, "Towards open data for linguistics: Lexical Linked Data," in *New Trends of Research in Ontologies and Lexical Resources*. Springer, 2013, pp. 7–25.
[5] J. P. McCrae, P. Cimiano, V. Rodríguez Doncel, D. Vila-Suero, J. Gracia, L. Matteis, R. Navigli, A. Abele, G. Vulcu, and P. Buitelaar, "Reconciling heterogeneous descriptions of language resources," in *Proceedings of the 4th Workshop on Linked Data in Linguisitcs*, 2015.

[6] J. P. McCrae and P. Cimiano, "Linghub: a Linked Data based portal supporting the discovery of language resources," in *Proceedings of the 11th International Conference on Semantic Systems*, 2015.

[7] J. P. McCrae, P. Labropoulou, J. Gracia, M. Villegas, V. R. Doncel, and P. Cimiano, "One ontology to bind them all: The META-SHARE OWL ontology for the interoperability of linguistic datasets on the Web," in *Proceedings of the 4th Workshop on the Multilingual Semantic Web*, 2015.

[8] G. Klyne, J. J. Carroll, and B. McBride, "RDF 1.1 concepts and abstract syntax," The World Wide Web Consortium, W3C Recommendation, 2006.

[9] C. Bizer, T. Heath, and T. Berners-Lee, "Linked data-the story so far," *Semantic Services, Interoperability and Web Applications: Emerging Concepts*, pp. 205–227, 2009.

[10] M. Nilsson, "From interoperability to harmonization in metadata standardization," Ph.D. dissertation, Royal Institute of Technology, 2010.

[11] M. Khoo and C. Hall, "Merging metadata: a sociotechnical study of crosswalking and interoperability," in *Proceedings of the 10th annual joint conference on Digital libraries*. ACM, 2010, pp. 361–364.

[12] J. Nogueras-Iso, F. J. Zarazaga-Soria, J. Lacasta, R. Béjar, and P. R. Muro-Medrano, "Metadata standard interoperability: application in the geographic information domain," *Computers, environment and urban systems*, vol. 28, no. 6, pp. 611–634, 2004.

[13] L. M. Chan and M. L. Zeng, "Metadata interoperability and standardization-a study of methodology, part i," *D-Lib Magazine*, vol. 12, no. 6, p. 3, 2006.

[14] J. Kunze and T. Baker, "The Dublin Core metadata element set," Internet Engineering Task Force, RFC 5013, 1997.

[15] C. Brooks and G. McCalla, "Towards flexible learning object metadata," *Continuing Engineering Education and Lifelong Learning*, vol. 16, no. 1/2, p. 50–63, 2006.

[16] M. Kemps-Snijders, M. Windhouwer, P. Wittenburg, and S. E. Wright, "Isocat: Corralling data categories in the wild." in *Proceedings of the 7th International Conference on Language Resources and Evaluation (LREC)*, 2008, pp. 887–891.

[17] K. Alexander, R. Cyganiak, M. Hausenblas, and J. Zhao, "Describing linked datasets with the VoID vocabulary," The World Wide Web Consortium, Tech. Rep., 2011, interest Group Note.

[18] F. Maali, J. Erickson, and P. Archer, "Data catalog vocabulary (DCAT)," The World Wide Web Consortium, W3C Recommendation, 2014.

[19] M. Brümmer, C. Baron, I. Ermilov, M. Freudenberg, D. Kontokostas, and S. Hellmann, "DataID: towards semantically rich metadata for complex datasets," in *Proceedings of the 10th International Conference on Semantic Systems*, 2014, pp. 84–91.

[20] C. Jenkins, M. Jackson, P. Burden, and J. Wallis, "Automatic rdf metadata generation for resource discovery," *Computer Networks*, vol. 31, no. 11, pp. 1305–1320, 1999.

[21] M. Gavrilidou, P. Labropoulou, E. Desipri, S. Piperidis, H. Papageorgiou, M. Monachini, F. Frontini, T. Declerck, G. Francopoulo, V. Arranz *et al.*, "The META-SHARE metadata schema for the description of language resources." in *Proceedings of the 8th International Conference on Language Resources and Evaluation*, 2012, pp. 1090–1097.

[22] S. Piperidis, "The META-SHARE language resources sharing infrastructure: Principles, challenges, solutions." in *Proceedings of the 8th International Conference on Language Resources and Evaluation*, 2012, pp. 36–42.

[23] H. v. d. Sompel, M. L. Nelson, C. Lagoze, and S. Warner, "Resource harvesting within the OAI-PMH framework," *D-Lib Magazine*, vol. 10, no. 12, 2004.

[24] D. Broeder, M. Windhouwer, D. Van Uytvanck, T. Goosen, and T. Trippel, "CMDI: a component metadata infrastructure," in *Describing LRs with metadata: towards flexibility and interoperability in the documentation of LR workshop programme*, 2012, p. 1.

[25] H. Tohyama, S. Kozawa, K. Uchimoto, S. Matsubara, and H. Isahara, "Shachi: A large scale metadata database of language resources," in *Proceedings of the 1st International Conference on Global Interoperability for Language resources*, 2008, pp. 205–212.

[26] K. Choukri, V. Arranz, O. Hamon, and J. Park, "Using the international standard language resource number: Practical and technical aspects." in *Proceedings of the 8th International Conference on Language Resources and Evaluation*, 2012, pp. 50–54.

[27] N. Calzolari, R. Del Gratta, G. Francopoulo, J. Mariani, F. Rubino, I. Russo, and C. Soria, "The LRE Map. Harmonising community descriptions of resources," in *Proceedings of the 8th International Conference on Language Resources and Evaluation*, 2012, pp. 1084–1089.

[28] C. Chiarcos, S. Hellmann, and S. Nordhoff, "The Open Linguistics Working Group of the Open Knowledge Foundation," in *Linked Data in Linguistics*. Springer, 2012, pp. 153–160.

[29] R. Del Gratta, G. Pardelli, and S. Goggi, "The lre map disclosed." in *Proceedings of the 9th International Conference on Language Resources and Evaluation*, 2014, pp. 3534–3541.

[30] A. Moro, A. Raganato, and R. Navigli, "Entity linking meets word sense disambiguation: a unified approach," *Transactions of the Association for Computational Linguistics (TACL)*, vol. 2, pp. 231–244, 2014.

[31] A. Seaborne, K. G. Clark, L. Feigenbaum, and E. Torres, "SPARQL 1.1 query results JSON format," The World Wide Web Consortium, W3C Recommendation, 2013.